%% file: mlsys-main.tex
\documentclass{article}

\usepackage{microtype}
\usepackage{graphicx}
\usepackage{subfigure}
\usepackage{booktabs} 

\usepackage{hyperref}

\usepackage{amsmath}

\usepackage[accepted]{mlsys2025}

\mlsystitlerunning{OSWorld-Human: Benchmarking the Efficiency of Computer-Use Agents}

\begin{document}
\input{macros}

\twocolumn[
\mlsystitle{OSWorld-Human: Benchmarking the Efficiency of Computer-Use Agents}




\mlsyssetsymbol{equal}{*}

\begin{mlsysauthorlist}
\mlsysauthor{Reyna Abhyankar}{equal,ucsd}
\mlsysauthor{Qi Qi}{equal,ucsd}
\mlsysauthor{Yiying Zhang}{ucsd,gensee}
\end{mlsysauthorlist}

\mlsysaffiliation{ucsd}{University of California, San Diego}
\mlsysaffiliation{gensee}{GenseeAI, Inc}

\mlsyscorrespondingauthor{Reyna Abhyankar}{vabhyank@ucsd.edu}
\mlsyscorrespondingauthor{Qi Qi}{qiqi@ucsd.edu}
\mlsyscorrespondingauthor{Yiying Zhang}{yiying@ucsd.edu}




\mlsyskeywords{Sys for ML, CUAs, LLM Benchmarks}

\vskip 0.3in

\input{abstract}
]



\printAffiliationsAndNotice{\mlsysEqualContribution} 

\input{intro-gemini}

\input{background-gemini}

\input{study-reyna}

\input{bench}

\input{evaluation}

\input{conclude}


\section*{Acknowledgements}
We would like to thank Zijian He and Vikranth Srivatsa for their valuable contributions and feedback on this paper.
This material is based upon work supported by gifts from AWS, Google, and Meta. Any opinions, findings, conclusions, or recommendations expressed in this material are those of the authors and do not necessarily reflect the views of these institutions.

\bibliography{references}
\bibliographystyle{mlsys2025}



\end{document}

%% file: macros.tex
%
%

\newcommand{\mm}{mm$^2$}
\newcommand{\figtitle}[1]{\textbf{#1}}
\newcommand{\us}{$\mu$s}
\newcommand{\fixme}[1]{{\color{red}\textbf{\fbox{FIXME} #1}}}
\newcommand{\FIXME}[1]{{\color{red}\textbf{\fbox{FIXME} #1}}}
\newcommand{\TODO}[1]{{\color{red}\textbf{\fbox{TODO} #1}}}
\newcommand{\NOTE}[1]{{\color{blue}\textbf{\fbox{NOTE} #1}}}
\newcommand{\note}[1]{{\color{blue}\textbf{\fbox{NOTE} #1}}}

\newcommand{\yiying}[1]{{\color{cyan}\textbf{\fbox{Yiying} #1}}}
\newcommand{\dongming}[1]{{\color{red}\textbf{\fbox{Yizhou} #1}}}
\newcommand{\pu}[1]{{\color{green}\textbf{\fbox{Ryan} #1}}}
\newcommand{\viks}[1]{{\color{olive}\textbf{\fbox{Will} #1}}}
\newcommand{\zijian}[1]{{\color{olive}\textbf{\fbox{zijian} #1}}}
\newcommand{\reyna}[1]{{\color{blue}\textbf{\fbox{Reyna} #1}}}



\newcommand{\myitem}[1]{\item \textbf{#1}}
\newcommand{\myitemit}[1]{\item \textit{#1}}

\newcommand{\x}{$\times$}
\newcommand{\eg}{\textit{e.g.}}
\newcommand{\ie}{\textit{i.e.}}

\newcommand{\sys}{OSWorld-Human}

%% file: abstract.tex
\begin{abstract}

Generative AI is being leveraged to solve a variety of computer-use tasks involving desktop applications. State-of-the-art systems have focused solely on improving accuracy on leading benchmarks. However, these systems are practically unusable due to extremely high end-to-end latency (\eg{} tens of minutes) for tasks that typically take humans just a few minutes to complete. To understand the cause behind this and to guide future developments of computer agents, we conduct the first study on the temporal performance of computer-use agents on OSWorld, the flagship benchmark in computer-use AI. We find that large model calls for planning, reflection, and judging account for most of the overall latency, and as an agent uses more steps to complete a task, each successive step can take 3\x{} longer than steps at the beginning of a task. We then construct \textbf{\sys}, a manually annotated version of the original OSWorld dataset that contains a human-determined trajectory for each task. We evaluate 16 agents on their efficiency using \sys\ and found that even the best agents take 2.7$-$4.3\x{} more steps than necessary.

\end{abstract}

%% file: intro-gemini.tex
\section{Introduction}

Computer-use agents, designed to autonomously control computer systems, have the potential to revolutionize productivity and accessibility. As generative AI (gen-AI) models grow increasingly powerful, these agents have surged in capability, now able to perform complex, multi-step tasks across a wide range of computer applications. Gen-AI models, particularly transformer-based large language models (LLMs)~\cite{transformer} and vision language models (VLMs)~\cite{dosovitskiy2020vit}, provide the core reasoning and perceptual abilities that allow agents to understand instructions, interpret screen content, plan actions, and execute them through keyboard and mouse inputs. Prominent examples include commercial products like OpenAI's Operator~\cite{openai_operator}, Anthropic's Claude Computer Use~\cite{claude}, Google's Project Mariner~\cite{mariner} and open-source projects like ByteDance's UI-TARS~\cite{ui-tars}, Agent S2~\cite{s2}, GTA1~\cite{gta1}, InfantAgent~\cite{infant}, and Jedi~\cite{jedi}.

Although these agents have shown impressive advancements in accuracy on complex benchmarks, a critical bottleneck remains: request latency. Our evaluation of real-world computer tasks suggests that computer-use agents can take tens of minutes to complete a task, which is in stark contrast to the couple of minutes a human expert might require for the same task. For instance, changing the line spacing of two paragraphs in a document to double-spaced takes 12 minutes for a computer-use agent. However, for a human user with introductory computer experience, this task should take under 30 seconds. The disparity limits the practical applicability of these agents, especially in interactive or time-sensitive scenarios, which hinders their integration into real-time workflows.

Prior research has predominantly focused on improving the accuracy and generality of computer-use agents, aiming to increase the percentage of tasks they can successfully complete. Although achieving high accuracy is always valuable, the temporal efficiency of these agents is equally crucial for real-world deployment and user experience. 

This paper presents the first in-depth study that analyzes the latency implications of computer-use agents. Specifically, we investigate the performance of state-of-the-art agents on the OS-World benchmark~\cite{osworld}, a realistic benchmark designed to evaluate multimodal agents in real computer environments (Ubuntu, Windows, MacOS) across a diverse suite of 369 tasks across 9 applications (Chromium~\cite{chromium}, GIMP~\cite{gimp}, LibreOffice Suite~\cite{libreoffice}, OS, Thunderbird~\cite{thunderbird}, VLC~\cite{chromium}, and Visual Studio Code~\cite{vscode}) via both graphical user interfaces (GUI) and command-line interfaces (CLI). 

To understand where latency originates within agent execution, we perform a detailed step breakdown analysis of agent trajectories on a representative set of 39 OS-World tasks using the Agent S2~\cite{s2} and GTA1~\cite{gta1} frameworks, two popular and leading open-source computer-use agents. We profile the steps taken by S2 and GTA1, including information retrieval, step planning, step grounding (\eg, finding coordinates), action-taking (\eg, mouse click, text input, keyboard shortcut), screenshotting, judging, and reflection. We find that the planning step accounts for more than half, sometimes close to 75\% of the total task latency, for both S2 and GTA1. The judging (GTA1) and reflection (S2) steps account for the second biggest time sink, accounting for 22.5\% and 33.6\% of total task time, respectively. Furthermore, these steps' latencies grow as an agent takes more steps to complete a task. 

To understand other factors affecting CUA's efficiency, we conduct token, failure, and cost analysis of GTA1. We find 23\% of errors are due to poor visual grounding, resulting in agents taking up to 30 extra steps on a single task. 

Overall, our study findings suggest three promising approaches for speeding up computer agent execution: (1) reducing the latency of planning, judging, and reflection calls, (2) minimizing the number of steps for a task, and (3) improving grounding or backoff mechanisms.

Based on this insight, we build \textbf{\textit{OSWorld-Human}}, a manually-constructed and set of human trajectories for all 369 OSWorld tasks. For each task, we meticulously annotate the minimal humanly-perceived steps required for successful task completion, based on verified ground-truth sources. Furthermore, we examine the actions that can be performed correctly from the same visual observation (screenshots). Knowing which actions can be ``grouped'' together has the implication that single planning, judgeing, and reflection steps are potentially sufficient for them, allowing future CUA systems to use much fewer LLM calls and achieving faster request latency.

We further propose a new efficiency-based metric, the \textbf{\textit{Weighted Efficiency Score (WES)}}, which penalizes inefficient trajectories for successful and failed tasks in separate ways. 
We apply this framework to anlayze the publicly-released trajectories of 16 popular computer-use agents against the single- and group-based manual step trajectories in OSWorld-Human. The best-performing agent with publicly-released trajectories on the OS-World leaderboard achieves a success rate of 41.4\% but only 15.6\% on \textbf{WES}. Our analysis using \sys\ shows that the leading agents still take 2.7\x{} to 4.3\x{} more steps than is required to complete the task. This stark difference highlights the need for more efficient and practical agents.



In summary, our work makes six key contributions:

\begin{itemize}
    \item The {\em first systematic study of temporal performance of CUAs} using two leading computer-use agents and the OSWorld benchmark.
    \item Detailed analysis of \textit{token counts, cost, observation types, and failures} in a step-by-step manner
    \item The identification of {\em LLM-based planning, reflection, and judging} steps to be the key bottlenecks in CUA request latency.
    \item The construction of \textit{OSWorld-Human benchmark}, a manually curated and cross-verified benchmark containing optimal trajectories for all 369 OSWorld tasks. \item The proposal of {\em grouping actions to reduce LLM calls}, and a human examined action grouping trajectories of all OSWorld tasks.
    \item \textit{New efficiency metrics and comprehensive evaluation} of 16 state-of-the-art agents, showing that even the best systems require sinificantly more steps than human trajectories.
\end{itemize}
Together, these contributions establish the first unified framework for benchmarking, analyzing, and improving the temporal efficiency of computer-use agents. 

\sys\ is available at \url{https://github.com/WukLab/osworld-human}.

%% file: background-gemini.tex
\section{Background and Motivation}
Autonomous agents capable of operating computer systems on behalf of human users represent a significant frontier in AI research. These computer-use agents (CUAs) aim to bridge the gap between high-level natural language instructions and low-level computer interactions, offering the potential to automate complex digital workflows and enhance accessibility. This section provides background on CUAs, their common technical approaches, and the latency challenge that motivates our work.

\subsection{Computer-Use Agents (CUAs)}
Computer-use agents are AI systems designed to perceive and interact with digital environments, such as operating systems, web browsers, and applications, much like a human user does. Their primary goal is to execute tasks by controlling the computer through standard interfaces, typically simulating keyboard inputs and mouse actions. 
CUAs can improve productivity by automating repetitive or tedious digital tasks, freeing up human users for more creative or strategic work. For individuals with disabilities, CUAs offer a potential pathway to increased computer accessibility by allowing interaction through natural language or other modalities. Furthermore, they serve as a challenging benchmark for evaluating the general capabilities of AI systems in complex, open-ended environments.

The rise of powerful large language models (LLMs) and vision language models (VLMs), such as GPT-4V~\cite{gpt-4v}, Qwen-VL~\cite{qwen-vl}, Llama-3~\cite{llama3}, and Gemini~\cite{gemini}, has significantly advanced CUA capabilities.
Several prominent research projects and emerging products demonstrate the progress in this area. OpenAI's Computer-Using Agent~\cite{openai_cua} is a notable example that leverages models like GPT-4o's~\cite{gpt-4o} vision capabilities to interpret raw screenshots to interact with the computer. 
Similarly, Anthropic has explored computer use capabilities, enabling models like Claude~\cite{claude} to interact with computer interfaces through defined tools and an agentic loop~\cite{anthropic_computer_use}. Other research focuses on refining perception and action spaces, such as Aguvis~\cite{aguvis}, which explores a pure vision-based framework for cross-platform GUI agents, OmniParser~\cite{omniparser}, which improves visual grounding through structured screen parsing, and UI-TARS~\cite{ui-tars}, a fine-tuned VLM for grounding tasks. GTA1~\cite{gta1} implements parallel rollouts and a judge to increase the likelihood of generating a valid action. Agentic systems like Agent S2~\cite{s2} and InfantAgent~\cite{infant} utilize a combination of retrieval and model-based methods to enhance computer use. 


\subsection{Common CUA Approaches}
\label{sec:cua-approahces}
Effective CUAs rely on sophisticated techniques for perceiving the environment and acting within it.

Perception is typically achieved through three primary modalities: computer \textit{screenshots}, which provides a direct visual representation of the current state to a model; \textit{accessibility trees}, which is a structured representation of UI elements that includes information about their roles, names, values, and hierarchical relationships, including unseen elements; and \textit{Set-of-Marks}~\cite{set_of_marks}, which overlays unique identifiers (marks) onto interactive UI elements in a screenshot and sends these marks to a model for prediction.

The action space defines the set of operations an agent can perform. For CUAs, this typically involves simulating fundamental human interactions like mouse movements, clicks (at specific coordinates and frequencies), scrolling, and keyboard inputs.

CUA architectures often combine perception and action generation within a planning and reasoning loop, typically powered by foundation and fine-tuned LLMs/VLMs using techniques like Chain-of-Thought~\cite{cot} or ReAct~\cite{react} to break down tasks, observe results, and correct errors.

\subsection{CUA Latency Challenges}
For CUAs to be seamlessly integrated into human workflows and to be useful for interactive tasks, their response time needs to approach human levels, especially in time-sensitive scenarios, such as rapid content editing, responding to frequent visual adjustments, and application execution. 

Reducing CUA latency is challenging due to several inherent factors. First, most computer tasks, and thus agent trajectories, involve a sequence of discrete steps. The total latency accumulates across all these steps, including perception, reasoning, and action execution for each step. Second, CUAs rely heavily on LLMs and VLMs. These models are called upon for planning, judging, and reflection at each step, and they often involve long prompts, further increasing their computation time. Third, unlike a human expert who often follows a direct, efficient path, CUAs may engage in trial-and-error, explore irrelevant parts of the interface, spend considerable time recovering from errors, or perform a correct but less efficient sequence of actions, all of which add significant latency.

Before latency inefficiencies can be addressed, a foundational step is to fully understand where inefficiencies come from and their severity for different computer-use tasks. Unfortunately, as far as we know, no existing work has properly studied the latency aspect of CUAs. This work directly investigates the impact of steps in CUA trajectories across a wide range of applications, provides a human-based gold standard, and directly identifies sources of inefficiency of CUAs.

%% file: study-reyna.tex
\section{The First CUA Latency Study}
\label{sec:study}

\input{fig_task_breakdown}

To understand the bottlenecks of computer-use agents, we study the performance of Agent S2~\cite{s2} and GTA1~\cite{gta1}, the leading open-source systems on the OSWorld leaderboard~\cite{osworld}. It should be noted that our analysis was conducted prior to the release of its successor, Agent S3.

\noindent\textbf{Methodology.}
We use Agent S2's default value that allows it to take up to 50 steps for each task. For GTA1, this value is 100 steps. The original S2 paper conducts their evaluation on both GPT-4o and Claude-3.7 as the foundational model for planning, reflection, and retrieval. In this study, we use GPT-4.1~\cite{gpt-4.1} as the planner, reflection, and retrieval models for S2. For GTA1, we used o3~\cite{openai2025o3o4mini} as the planner and judge, as described in their paper. For grounding, we use UI-TARS-7B-DPO~\cite{ui-tars} and GTA1-7B~\cite{gta1} for S2 and GTA1, respectively. The grounding model is hosted on a single NVIDIA A6000 GPU using SGLang~\cite{sglang}, a state-of-art inference engine. We utilize the OSWorld-provided subset of 39 tasks, or 10\% of the entire benchmark. We then run Agent S2 and GTA1 on each of the tasks and collect detailed timing and token traces. 

 
\noindent\textbf{S2 Background.} 
At the beginning of each task, the S2 agent performs a one-time retrieval step. During this stage, the agent calls the large language model (LLM) to retrieve the most similar narrative or prior task experience from a local database and, if a \texttt{search\_engine} is provided, additionally retrieves external knowledge from the web. The retrieved information is incorporated into the initial prompt to form a high-level task plan. After this initialization, S2 proceeds through an iterative perception–planning–execution loop. At each step, the agent first captures a screenshot of the current environment (non-LLM). It then calls the planner model (LLM) to generate a per-step plan based on the screenshot and the accumulated task history. Next, the grounding model (a smaller vision model) translates this textual plan into concrete screen coordinates, after which the agent executes the corresponding action (mouse click, text input, or key press) directly in the environment (non-LLM). After execution, the agent calls the reflection model (LLM) to analyze the resulting screenshot and determine whether the previous action achieved its intended effect, informing the next planning step. This process repeats until the task is completed, deemed infeasible, or reaches the maximum step limit. Overall, S2 involves three major LLM-based operations—retrieval (once), planning (per step), and reflection (per step)—while grounding and execution are handled by smaller models or direct system actions.

\noindent\textbf{GTA1 Background.} 
In contrast, GTA1 omits the initial retrieval phase and relies on a parallel planning and judging mechanism. At each step, the agent first captures the current screenshot (non-LLM) and then sends it to the planner model (LLM) through parallel calls, producing multiple candidate actions. These candidate plans are then evaluated and aggregated by a judge model (LLM), which selects a single action to execute. The selected action is passed to the grounding model (a smaller vision-language model) that converts it into precise coordinates, and the resulting action is executed directly in the environment (non-LLM), producing a new screenshot for the next iteration. GTA1 does not include a reflection stage; instead, it relies on its repeated judging mechanism to implicitly correct suboptimal plans over time. The loop continues until the task is completed, fails, or reaches the predefined step limit. As a result, the dominant LLM calls in GTA1 occur during the parallel planning and judging phases, while grounding and execution remain lightweight operations.

\subsection{Task Latency Analysis}\label{an-latency}

\input{fig_cdf_comparison_grid}

Figure \ref{fig-task-breakdown} illustrates a timeline view of how S2 and GTA1 completes a task in the OS domain. Specifically, the task is to create an SSH user with a given username and password that is only allowed access to a specific folder. The top row shows S2 completes this task in 50 steps, which is the maximum number of steps allowed. Planning, reflection, and retrieval all leverage the large foundation model (in this case, GPT-4.1). From the figure, per-step planning and reflection is responsible for the bulk of the end-to-end latency of over 40 minutes. The bottom row shows the timeline breakdown of GTA1 on the same task, following the workflow of planning, judging, grounding and executing. Similar to Agent S2, GTA1 completes this task successfully in 54 steps, whereas the end-to-end latency is almost twice that of Agent S2. The parallel rollout planning phase and subsequent repetitive judging phase of GTA1 were the main reasons for its extended duration.

\input{tab_task_breakdown_by_agent}

We further break down the average time spent in each stage by application in Table \ref{tab:agent_time_breakdown}. For Agent S2, planning and reflection makes up 76\% to 96\% of the total task latency. These tasks involve LLM/VLM calls usually with long context (thousands of tokens), explaining their long latency. Planning takes more time than reflection overall because planning takes place at each step and after the completion of a sub-task, while reflection only takes place at each step. Retrieval accounts for 0.7\% to 8.9\% of the overall latency. While retrieval does leverage the large model, it only occurs once at the beginning of the task, which means its overhead is constant. Meanwhile, grounding invokes a much smaller model and is served using a popular and efficient inference engine, SGLang~\cite{sglang}. The throughput of the grounding model is dependent on both the size of the model, the performance of the GPU, and the serving engine. 

GTA1 exhibits a similar phenomenon, with planning and judging accounting for 91-96\% of total task latency. The planner model retains the ability to retry 3 times if no valid plan within the action space is generated. Meanwhile, screenshot and action execution are the least intensive for both agents since neither of them require significant GPU resources like large models do. It is clear that gen-AI model calling dominates the latency for a task, hence we conduct a more thorough per-step breakdown.

\subsection{Model Call Analysis}

\input{fig_latency_by_step}
\input{fig_gta1_latency_by_step}

The above latency analysis shows that large model calls are the major contribution to task end-to-end latency.
To understand LLM/VLM calls in CUAs, we analyze them at different steps in a trajectory of Agent S2 for each task.
Figure \ref{fig-latency-step} and Figure \ref{fig-gta1-latency-step} shows the call latency and prompt token count distributions of Agent S2 and GTA1 respectively across all our subset of 39 tasks. Each line represents the average value across five consecutive steps (ten lines in total for 50 maximum steps for Agent S2 and 100 maximum steps for GTA1). In each figure, when getting to the later steps, large model calls take longer and have longer prompts. This is due to the prompting mechanism for most CUAs: at each step, the prompt sent to the LLM includes the history of all previous steps. For example, if the agent is on step 10, the prompt will include the screenshot for steps 1-9, along with planning details and reflection feedback. With double the step limit and parallel planning calls, GTA1 achieves higher overall accuracy at the cost of a larger prompt token range (28.5\% higher) and higher latency per step. 




We further study the comparison across different types of LLM calls (planning, reflection, retrieval, and grounding for Agent S2; planning, judging, and grounding for GTA1). Figure~\ref{fig-cdf-comparison-grid} plots the distribution of prompt tokens, completion tokens, and the latency of each type of LLM calls in a CDF. The upper row shows in Agent S2 grounding maintains a constant and small number of prompt and output tokens because it only receives a single screenshot, as opposed to the entire history being sent to planning, reflection, and retrieval models. Furthermore, the output of the grounding model is a set of coordinates, which is always fixed at 12 tokens in this framework. The bottom row plots the same metrics for GTA1, showing a similar pattern, but its distribution is smoother due to GTA1's parallel planning and multiple judging attempts at each step. 

The latency of planning, reflection, and judging is dominated by the prefill stage of LLM inference due to the large number of prompt tokens. Retrieval, however, does not take a screenshot as input since it occurs at the beginning of the task. Instead, the LLM only processes the relevant document retrieved from the database and the user's question. Therefore, it has a stable and relatively small number of prompt tokens. Instead, its latency is dominated by the decoding stage since the model is tasked with integrating the document into the execution plan. Since retrieval only occurs once, its overall contribution to the end-to-end latency of a task is minimal.




\subsection{Observation Types}

In addition to studying LLM call behavior, we also study how different approaches to perception, specifically screenshot, accessibility (A11y) trees, and Set-of-Marks (SoM)~\cite{set_of_marks} (\S\ref{sec:cua-approahces}), affect task latency. We selected Agent S2 for this analysis because it is representative in its thorough support for all three modalities, a feature not present in all CUAs (e.g., GTA1 only uses screenshots).

\input{fig_a11y_bar}

We sampled one task per application (excluding multi-app workflows) and charted task latency in Figure \ref{fig-a11y-duration} with a 10k token cutoff for the A11y tree. SoM does not incur significant overhead from labeling the screenshot and hence is omitted from figure \ref{fig-a11y-duration}. By and large, inclusion of the A11y tree drastically increases the per-task latency for two reasons, though this effect varies based on the application. First, generating the tree itself takes time and is dependent on the elements that exist in the current window, including hidden and visible elements. This can take anywhere from 3 seconds to 26 seconds. Applications that have more elements, such as the LibreOffice suite~\cite{libreoffice}, incur a much higher latency from generating the tree. Other applications, such as GIMP~\cite{gimp}, may benefit from utilizing the tree to complete the task quicker. Second, the tokenized tree is included in each prompt to the model, which can be thousands more tokens per step. This can significantly affect tasks with longer trajectories. 

\input{tab_obs_type_num_steps_comp}
We also show the number of steps taken for each task across observation types in Table \ref{tab:obs-type-step-comp}. Adding A11y trees to screenshots (SS) increases the number of steps for most applications. Visually rich applications like the LibreOffice Suite~\cite{libreoffice} see a particularly high increase because trees can contain thousands of nodes for such applications. Adding A11y tree decreases the number of steps for OS, GIMP~\cite{gimp}, and Chrome~\cite{chromium}, because either the application contains fewer distinct visual elements or the tree assists the CUA in completing the task faster. Adding SoM in addition to A11y trees lowers the number of steps overall and achieves the fewest number of steps for LibreOffice Writer, VLC~\cite{vlc}, Thunderbird~\cite{thunderbird}, and GIMP. It does, however, use more steps for Visual Studio Code~\cite{vscode} and Chrome~\cite{chromium} while SoM ties the SS+A11y result in the highest steps for LibreOffice Impress. This is likely task-dependent, as opposed to application-dependent. Overall, the marked screenshot contains useful information for the model to complete the task using fewer steps.

\subsection{Detailed Cost and Failure Analysis}
\label{sec:cost-and-failure}
While section \ref{an-latency} showed that both Agent S2 and GTA1 share a similar overall structure where LLM calls are the primary latency bottleneck, GTA1's architecture is more complex. Its use of parallel planning rollouts and a subsequent judging step is computationally intensive and costly, intended to improve plan quality. This complexity makes it a more salient case for a deeper investigation into compounding inefficiencies. We therefore focus our detailed cost and failure analysis on GTA1, as it serves as a representative upper bound for the challenges in state-of-the-art CUA systems.

\noindent\textbf{Cost Analysis}
GTA1 makes multiple planning model calls to improve the plan quality. Although effective, a drawback is the increased dollar cost. To understand this tradeoff, we perform a cost analysis of GTA1. For GTA1, its planner consists of two stages, planning and judging. Table~\ref{tab:cost_analysis} shows the total cost and prompt count breakdown across different steps (across all 39 tasks). On average, a task costs \$2.43, with planning, judging, and grounding being responsible for 87\%, 13\%, and less than 1\%, respectively. Planning costs over 6\x{} more than judging because in each planning step, GTA1 makes 4 parallel calls to o3. If any calls fail to generate a valid plan, they will retry that call up to 3 times. Hence, for every judging call, there can be between 4 and 12 planning calls.  Figure \ref{fig-gta1-cost} shows accumulated cost increases over steps. The quadratic increase contributes to the drastic cost difference between the planning and judging models compared to the grounding model. 

\input{tab_cost_analysis}

\input{fig_gta1_cost_analysis}

\noindent\textbf{Failure Analysis.}
For GTA1 specifically, the grounding model fails to efficiently locate the coordinates to execute the action, which leads to tens more steps for the same generated plan. For instance, a single action of opening a folder in Visual Studio Code takes 100 steps to fail; while the planner generated the correct action (``Click the blue `Open' button''), the grounding model was unable to generate precise coordinates. This resulted in the agent repeating the same step, accounting for 10\% of all planning steps. In fact, 23\% of all failures for GTA1 exhibited such repetitive behavior, where the planning step is correct but the grounding model generates incorrect coordinates. There was also no efficient backoff mechanism. When a wrong step is taken, the planning model would take \textit{several more} steps to realize that a wrong path has been taken. Instead of jumping directly back to the last step that was correct, the agent tries to perform new actions to go back to a previous state. 


For example, Chrome task \texttt{bb5e4c0d} asks the agent to ``set Bing as default search engine." The agent fails by reaching the 100-step limit. There are 10 loops accounting for a total of 72 wasted steps. The worst offender is the loop from steps 23-41. The planner correctly intends to ``click the search engine list in settings and scroll to find Bing", but the grounding model repeatedly outputted coordinates locked around the tab bar instead of the settings pane.
The planner's reflections in successive steps repeatedly noted ``still showing the same engines", yet the grounding model never corrected. After 18 such steps, the planner abandoned its intention. This loop costed \$8.47 and 27 minutes of wall-clock time.

We further analyzed tasks that failed with more than 50 steps. To detect whether the agent is in a loop due to grounding errors, we enforce that step $i$ must share at least 2 of the following criteria with step $i+1$: similarity of text generated by the planner, similarity of the action text, coordinate clustering within 50px, and screenshot perceptual hash match within 5\% of pixels. Based on this criteria, we detect that 66\% of steps are wasted by the agent being stuck in a loop due to grounding errors. Furthermore, the agent only recovers 54\% of the time; in other cases, it either abandons its approach entirely (37\%) or reaches the step limit (9\%).

When adjusting Table \ref{tab:agent_time_breakdown} for steps caused by grounding failures, we see planning go down from 75\% to 67\%, judging go down from 23\% to 21\%, and grounding go up from 2\% to 12\%.

\subsection{Study Generalizability}

While our study focuses on Agent S2 and GTA1, its implications go beyond a single agent. This is because both agents share the same general framework, and it is representative of other agentic systems. In each step, Agent S2 and GTA1 call a series of models with the observation as part of the prompt to output the final action(s) for a step. While the exact model calls and prompts may differ, this pattern is common in other CUA systems. For example, InfantAgent~\cite{infant}, a top-5 solution on the OSWorld leaderboard as of May 2025, uses the same framework: it invokes reasoning and utilizes previous history to output a set of actions. Then, it performs an evaluation (\ie{} judgement) and summary (\ie{} reflection) before starting the next step. Another system, Jedi~\cite{jedi}, uses two models as well: one large model for planning (\eg{} GPT-4o) and a fine-tuned smaller model for grounding, while following the same iterative pattern. 

A more straightforward alternative to using agentic systems is to make a single call to a fine-tuned model, such as the UI-TARS-1.5~\cite{ui-tars}. This study is still applicable to calling a single model for two reasons. First, the framework of ``observe-call-act" remains the same. Hence, the behavior of a single call to a large model can resemble an agentic system that deploys multiple small models, both in latency and tokens consumed. Second, the overall trend is towards agentic systems in these settings due to their superior performance. When Claude 3.7 was released in February 2025, it occupied the top spot on the OSWorld leaderboard with a success rate of 28\% using 100 steps. Three weeks later, Agent S2 with Claude 3.7 outperformed the original model with a score of 34.5\% in only 50 steps. 







%% file: fig_task_breakdown.tex
\begin{figure*}[ht!]
  \begin{center}
  \includegraphics[width=\textwidth]{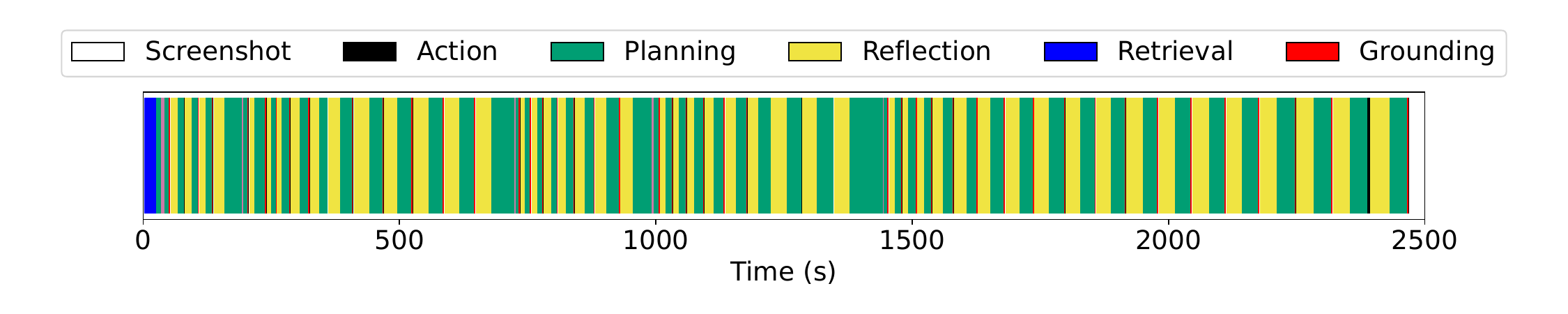}
  \includegraphics[width=\textwidth]{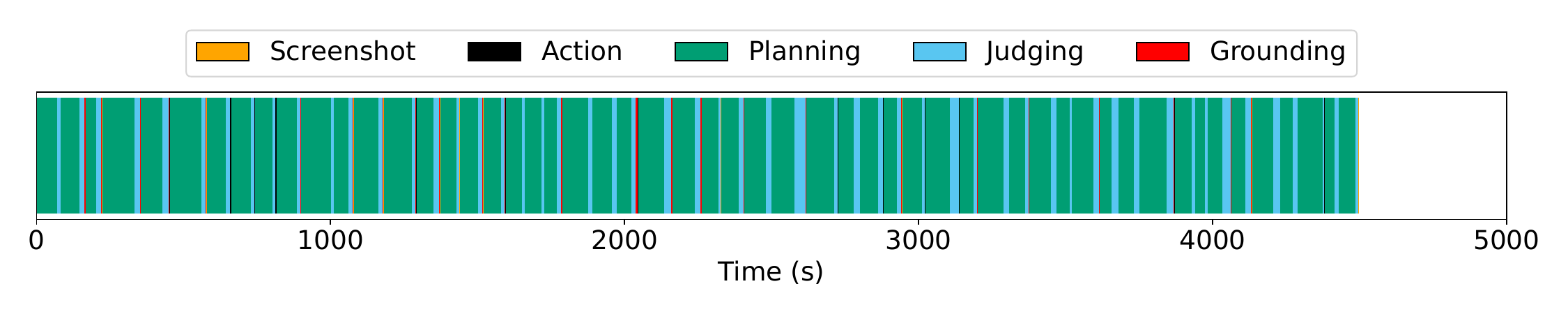}
  \caption{Timeline view of an OS task successfully completed by Agent S2 (top row) and GTA1 (bottom row).}
  \label{fig-task-breakdown}
  \end{center}
\end{figure*}

%% file: fig_cdf_comparison_grid.tex
\begin{figure*}
  \begin{center}
  \includegraphics[width=0.8\textwidth]{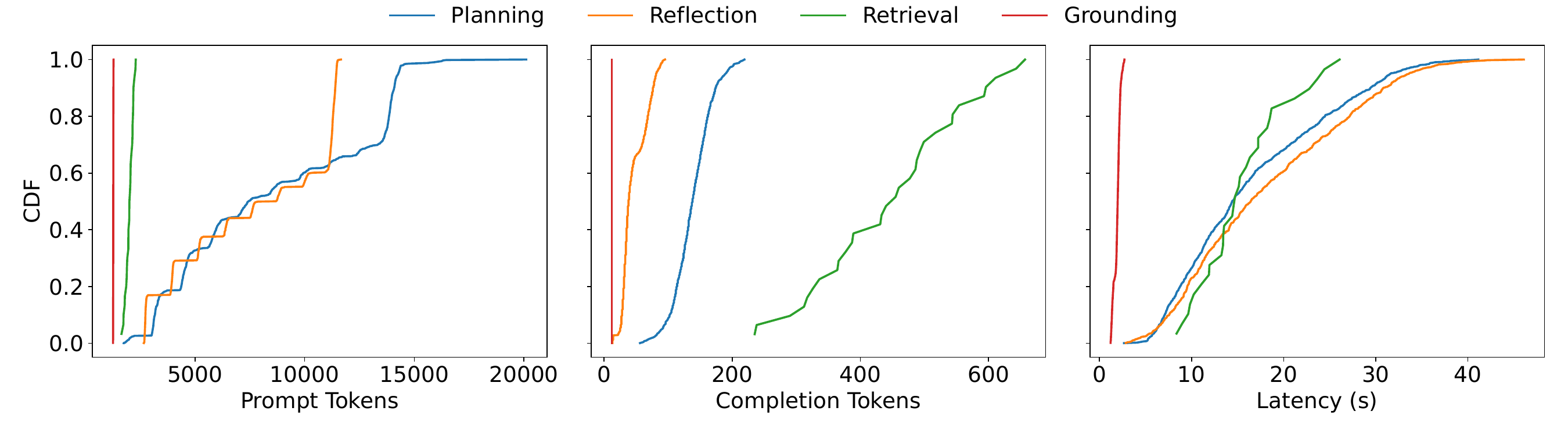}
  \includegraphics[width=0.8\textwidth]{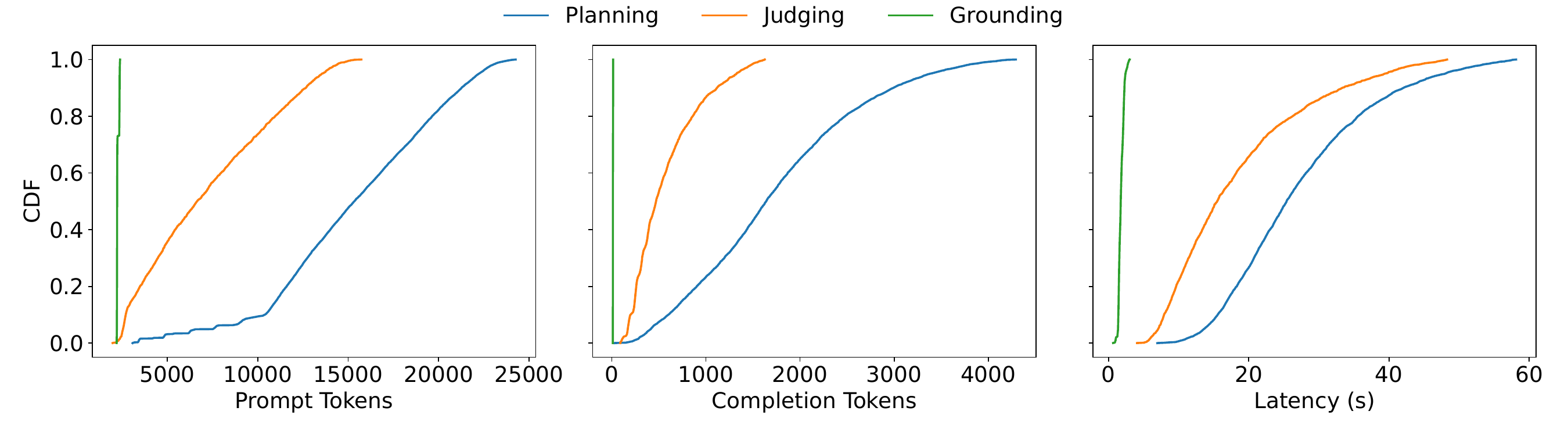}
  \caption{Comparison of prompt tokens, completion tokens, and latency for different types of LLM calls for Agent S2 (top row) and GTA1 (bottom row).}
  \label{fig-cdf-comparison-grid}
  \end{center}
\end{figure*}

%% file: tab_task_breakdown_by_agent.tex
\begin{table*}
\small
\begin{center}
\caption{Average percentage breakdown of time spent performing a certain sub-task grouped by agent.}
\begin{tabular}{|l|*{7}{c|}}
\hline
\textbf{Agent} & \textbf{Screenshot} & \textbf{Action} & \textbf{Planning} & \textbf{Judging} & \textbf{Reflection} & \textbf{Retrieval} & \textbf{Grounding} \\
\hline
GTA1 & $0.36 \pm 0.35$\% & $0.7 \pm 0.36$\% & $74.59 \pm 2.55$\% & $22.53 \pm 2.85$\% & - & - & $1.82 \pm 0.84$\% \\
S2 & $1.72 \pm 0.73$\% & $1.61 \pm 0.35$\% & $53.47 \pm 3.18$\% & - & $33.56 \pm 7.77\%$ & $3.08 \pm 2.14\%$ & $3.91 \pm 0.83$\% \\
\hline
\end{tabular}
\label{tab:agent_time_breakdown}
\end{center}
\end{table*}

%% file: fig_latency_by_step.tex
\begin{figure}[htb]
  \centering
  \includegraphics[width=\columnwidth]{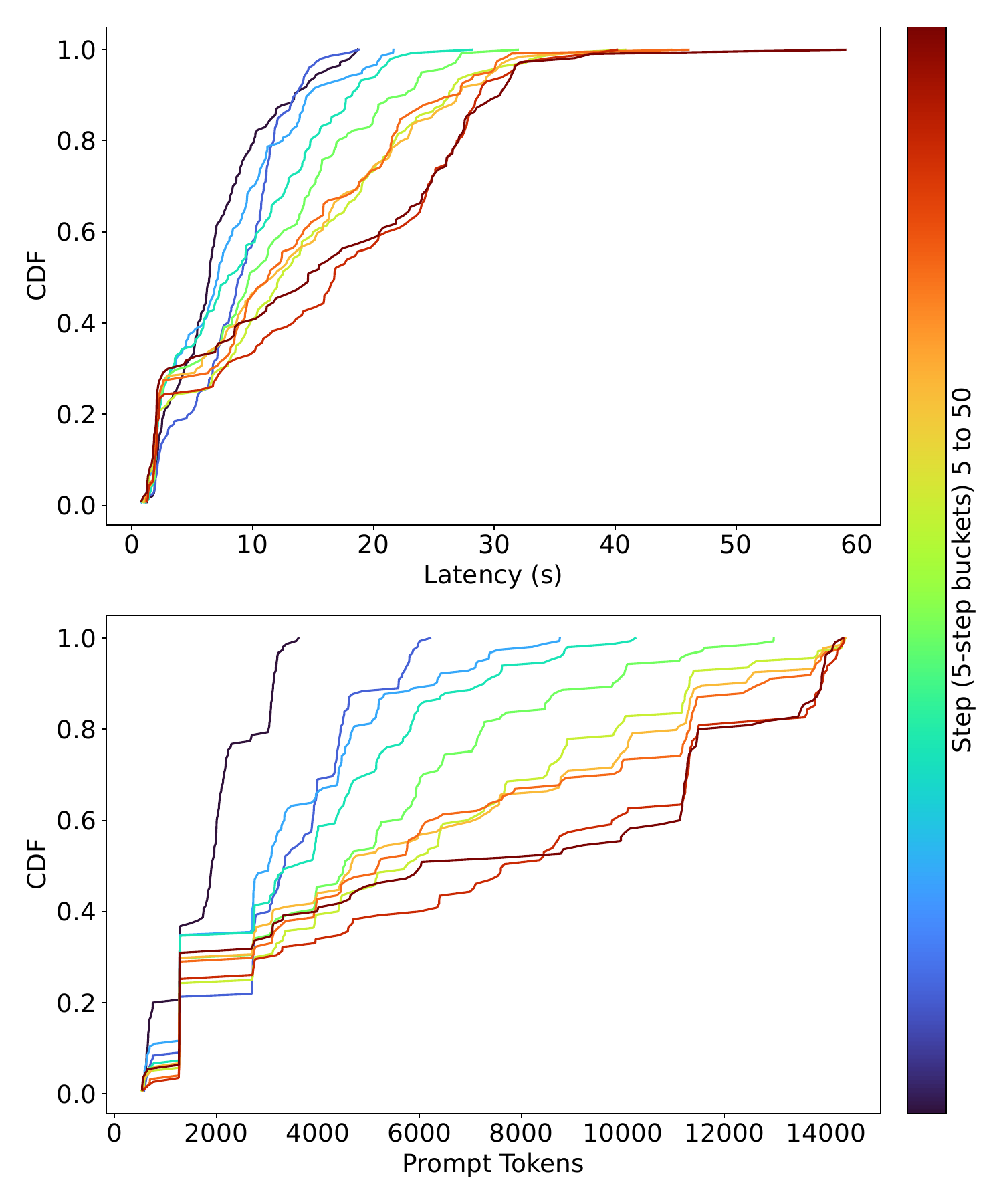}
  \caption{Probability distribution of latency and prompt tokens for a single step.}
  \label{fig-latency-step}
\end{figure}

%% file: fig_gta1_latency_by_step.tex
\begin{figure}[htb]
  \centering
  \includegraphics[width=\columnwidth]{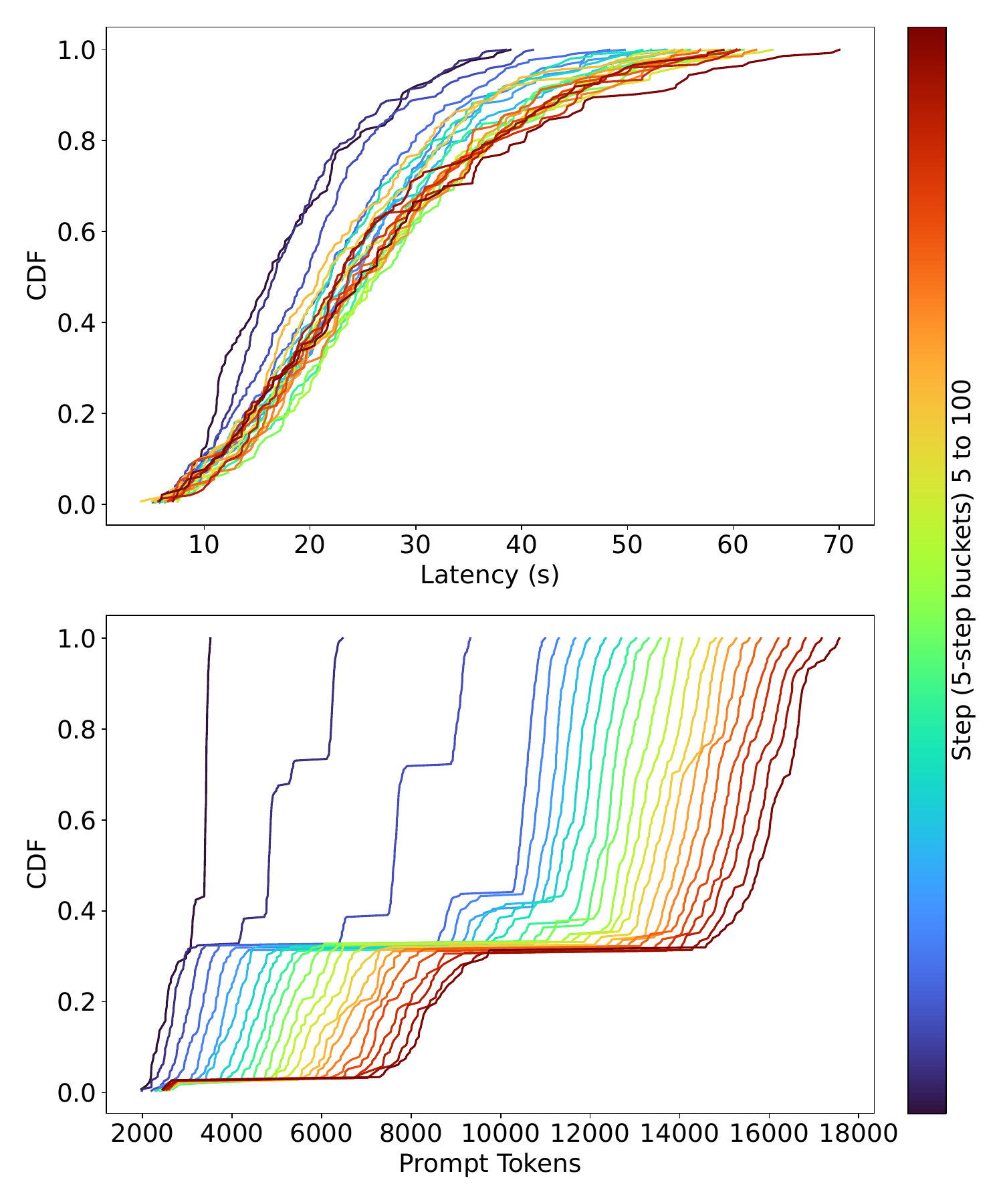}
  \caption{Probability distribution of latency and prompt tokens for a single step (GTA1).}
  \label{fig-gta1-latency-step}
\end{figure}

%% file: fig_a11y_bar.tex
\begin{figure}[htb]
  \centering
  \includegraphics[width=\columnwidth]{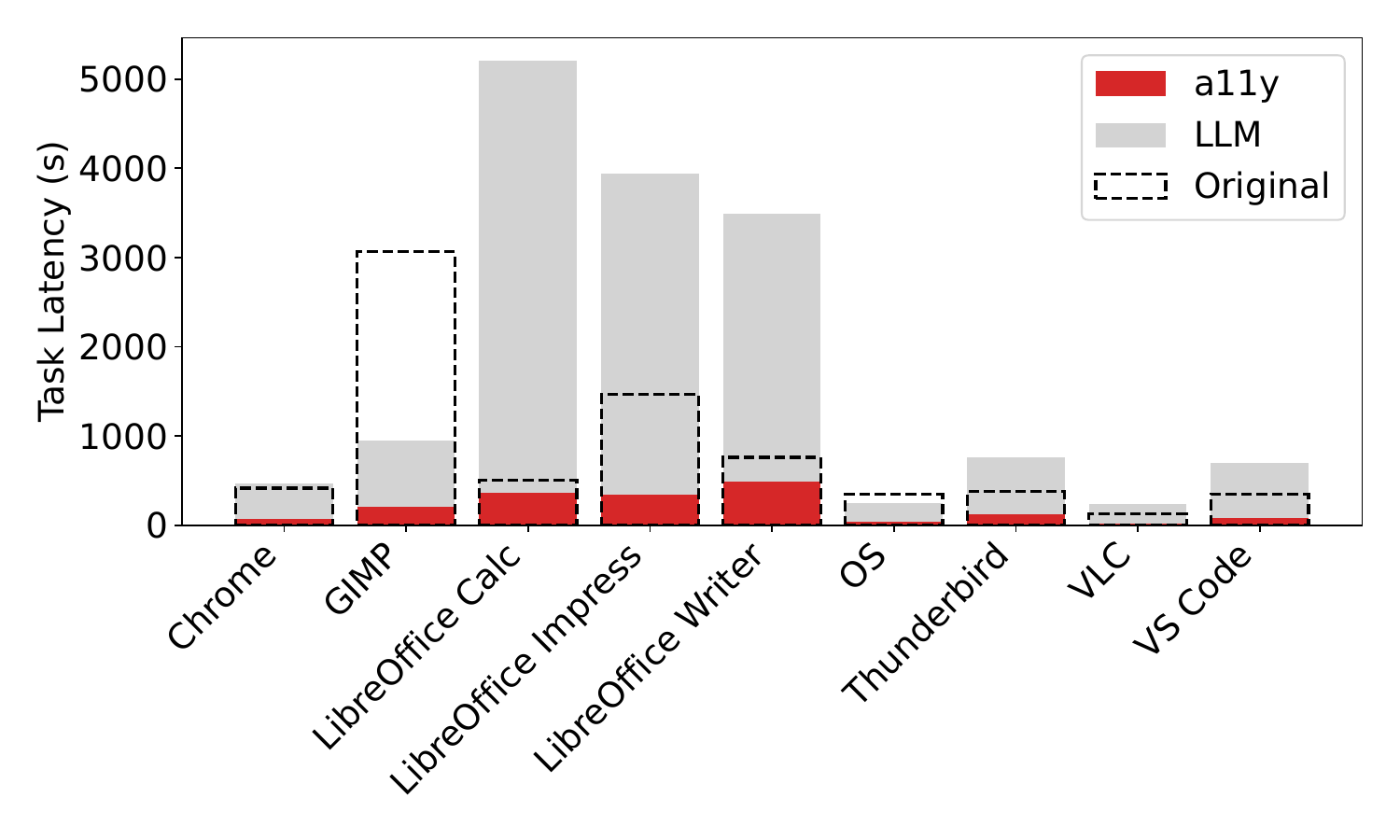}\vspace{-1.5em}
  \caption{End-to-end task latency for each application. Dashed box represents screenshot-only latency while colored bars show the breakdown when including the a11y tree.}
  \label{fig-a11y-duration}
\end{figure}

%% file: tab_obs_type_num_steps_comp.tex
\begin{table}[h!]
\small
  \centering
  \caption{Number of steps per sampled task per application for each observation type. Bolded highlights the fewest steps for that application's task.}
    \vspace{1em}
  \begin{tabular}{|l|c|c|c|}
    \hline
    \textbf{Application} & \textbf{SS} & \textbf{SS+A11y} & \textbf{SS+A11y+SoM} \\
    \hline
    OS & 8 & \textbf{5} & \textbf{5} \\
    Thunderbird & 14 & 20 & \textbf{8} \\
    VS Code & \textbf{9} & 13 & 16 \\
    LibreOffice Writer & 21 & 50 & \textbf{16} \\
    VLC & 3 & 3 & \textbf{2} \\
    GIMP & 49 & 17 & \textbf{12} \\
    LibreOffice Impress & \textbf{32} & 50 & 50 \\
    Chrome & 11 & \textbf{8} & 17 \\
    LibreOffice Calc & \textbf{14} & 50 & 10 \\
    \hline
  \end{tabular}
  \label{tab:obs-type-step-comp}
\end{table}

%% file: tab_cost_analysis.tex
\begin{table*}
\small
\begin{center}
\caption{Cost Analysis of Different Types of LLM Calls. Unit price using billing from TogetherAI~\cite{togetherai_pricing}}
\begin{tabular}{|l|l|l|l|l|l|l|}
\cline{1-7}
\textbf{Model}           & \textbf{Prompt Tokens} & \textbf{Prompt Price} & \textbf{Prompt Cost} & \textbf{Output Tokens} & \textbf{Output Price} & \textbf{Output Cost} \\ \hline
Planning (o3) & 87.34$\times 10^6$        & \$2 per mil     & \$174.69              & 10.29$\times 10^6$           & \$8 per mil  & \$82.35              \\ \hline
Judging (o3) & 17.94$\times 10^6$        & \$2 per mil     & \$35.89              & 1.59$\times 10^6$           & \$8 per mil  & \$12.76              \\ \hline
Grounding (GTA1-7B)    & 3.77$\times 10^6$        & \$0.30 per mil   & \$1.13                & 0.02$\times 10^6$           & \$0.30 per mil   & \$0.006               \\ \hline
\end{tabular}
\label{tab:cost_analysis}
\end{center}
\end{table*}

%% file: fig_gta1_cost_analysis.tex
\begin{figure}[htb]
  \centering
  \includegraphics[width=\columnwidth]{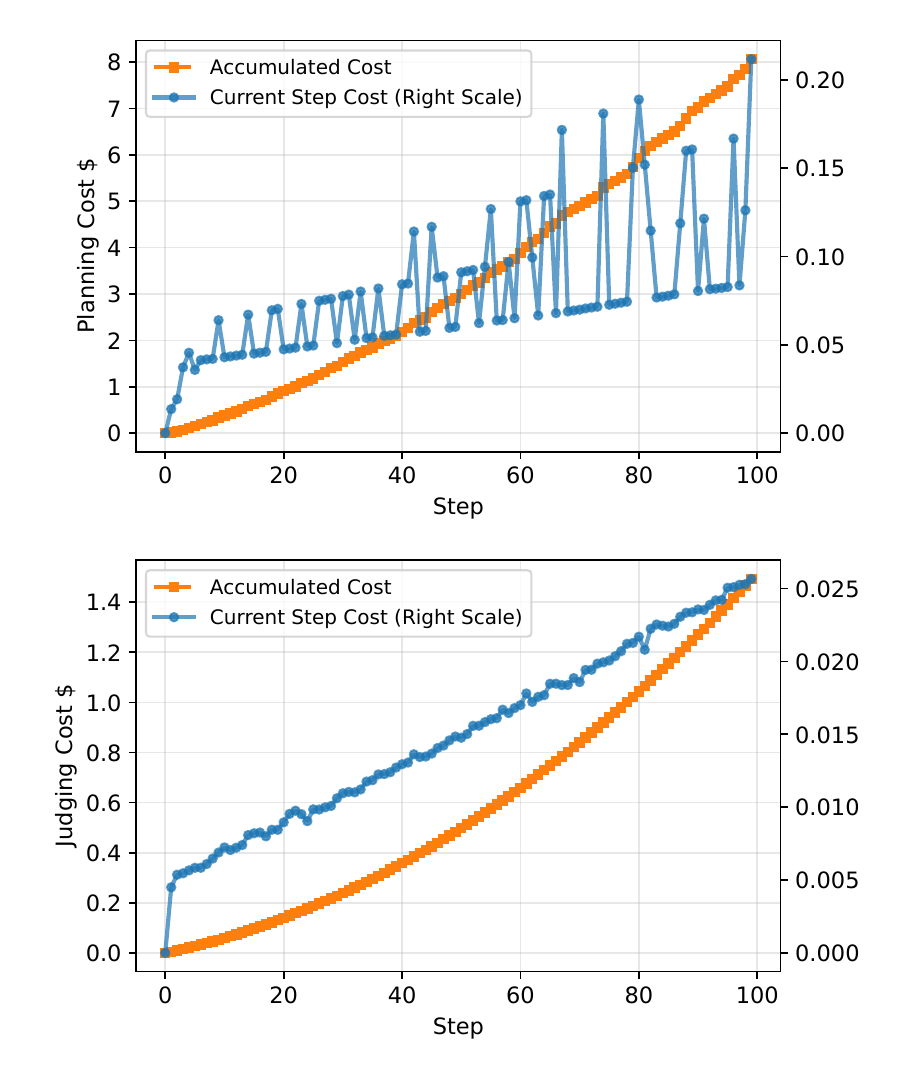}
  \caption{Accumulated and per-step cost for GTA1 on a VS Code task.}
  \label{fig-gta1-cost}
\end{figure}

%% file: bench.tex
\section{\sys}
\label{sec:gold}

Based on the findings in our study, we constructed \sys, a manually annotated version of OSWorld that lists the minimal humanly-perceived steps required to successfully complete a task. We now describe our construction of \sys.

\input{tab_single_vs_multi_per_app}

\subsection{Human Trajectory Construction}

To construct \sys, we first seek, perform, and verify ground-truth trajectories (\ie, steps) for all OSWorld tasks.
In the original OSWorld benchmark, most tasks contain a ``source'' that details a concrete ground-truth trajectory for solving the task. For example, users may ask a question in an online forum regarding editing their browser settings. The responses given in the forum provide the set of actions needed to complete the task. We manually map the steps given by the source to the action space defined in OSWorld and verify its success in the virtual environment. We refrain from listing exact coordinates, as these may change across environments.

For tasks without a listed source or where the source is ambiguous, we manually identify the necessary steps and compare with the ground truth from the original benchmark (\eg, OSWorld provides a gold file for all LibreOffice tasks). For all applications besides OS, we refrain from using any programmatic methods (unless mentioned in the source) to solve the task (for example, using the LibreOffice scripting language to modify a given spreadsheet), as that is beyond the expectation of what the typical user can do. However, we do utilize application-specific information, such as keyboard shortcuts, as a typical user is likely to be familiar with them. 

The dataset was constructed using two passes by two computer science graduate students. Then, each student cross-validated the other's results for a consensus. The final dataset was validated by manually performing the actions in the OSWorld virtual machine setup and obtaining a successful evaluation score for each task. \sys\ contains the same number of examples (369) as the original OSWorld.

\subsection{Action Grouping}
\label{sec:action-grouping}

When inspecting ground-truth actions taken to accomplish OSWorld tasks, we find that multiple actions can often be performed consecutively in sequence without any change in observation or prediction in between. For example, the following steps can be performed in succession without the need for more detailed planning: click on a text field, type text, and press $\texttt{<enter>}$.
This implies that CUA systems can potentially perform one observation and model call for a \textit{group} of actions, thereby reducing the number of steps and overall task latency.

To provide insights into this feature, we construct a \textbf{single-action} trajectory as well as a \textbf{grouped-action} trajectory for each task in \sys. The single-action trajectory lists all actions necessary to complete a task. 
The grouped-action trajectory consists of groups of actions that can be correctly executed from the same visual observation. For example, a list of actions may include clicking on a cell, typing a formula, and filling the column. These actions can be grouped together because the UI elements (and the coordinates) needed for all actions are present in a single screenshot. The grouped-action trajectory contains strictly less than or the same number of steps as the single-action trajectory.

Table~\ref{tab:app-single-multi} summarizes our construction of single- and grouped-action trajectories. For grouped-action trajectories, we count each group as a step, while single-action trajectories use one step per action. Overall, all applications benefit from grouping; the total number of steps and thus large model calls are reduced.
For applications where actions frequently trigger new windows, popups, or otherwise significant UI changes, there is a smaller difference between single-action and grouped-action trajectories. On the other hand, applications that operate on the same page or sheet, such as LibreOffice Writer or LibreOffice Calc, are much less disruptive.

%% file: tab_single_vs_multi_per_app.tex
\begin{table}[h!]
\small
  \centering
  
  \caption{Average Steps per Trajectory by Application}
    \vspace{1em}
  \begin{tabular}{|l|c|c|}
    \hline
    \textbf{Application} & \textbf{Single} & \textbf{Grouped} \\
    \hline
    OS & 3.9 & 2.0 \\
    Thunderbird & 6.7 & 3.8 \\
    VS Code & 3.6 & 2.0 \\
    LibreOffice Writer & 7.5 & 3.2 \\
    VLC & 5.1 & 3.7 \\
    GIMP & 2.8 & 2.0 \\
    LibreOffice Impress & 7.8 & 4.0 \\
    Chrome & 5.8 & 4.3 \\
    LibreOffice Calc & 13.2 & 4.5 \\
    \hline
  \end{tabular}
  \label{tab:app-single-multi}
\end{table}

%% file: evaluation.tex
\section{Evaluation}

\input{tab_full_results_update}

This section presents our metric and evaluation of 16 leading CUAs with published trajectories on the OSWorld leaderboard at the time of writing, conducted on all 369 examples in \sys.

\subsection{Weighted Efficiency Score}

As \sys's goal is to evaluate CUA systems' temporal performance, we want to approximate end-to-end latency by measuring how closely an agent performs compared to ground-truth human trajectories.
An easy way to measure this is to compare the minimum expected, human-performed number of steps across all possible trajectories ($t_{human}$) to the actual, agent-generated number of steps ($t_{agent}$) for a task. However, \textit{only} looking at efficiency favors an agent that fails a task with fewer steps over one that succeeds with more steps.

To accommodate this issue, we propose a new metric for evaluating CUAs: \textbf{Weighted Efficiency Score}, or \textbf{WES}. For a task $t$ that was completed successfully ($r_t = 1$), we weight the result based on its efficiency (\ie{} ${t_{human}} / {t_{agent}}$). This is based on the premise that a success in fewer steps is strictly preferable to a success that takes more steps. For a task $t$ that was unsuccessful ($r_t = 0$), this efficiency term is zero. We compute the per-task efficiency over all $n$ tasks in the dataset to obtain $\text{WES}^+$:

\begin{align*}
    \text{WES}^+ &= \frac{1}{n}\sum_t^n r_t \cdot \frac{t_{human}}{t_{agent}}
\end{align*}

$\text{WES}^+$ typically ranges from 0 to 1, though it is possible for an agent to complete a task faster than a human (\ie{} by leveraging its latent capabilities instead of performing actions on-screen). A score of 0 can mean an agent fails to complete most tasks or it is very inefficient. An agent that scores closer to 1 is both successful and efficient.

To penalize agents that waste steps on failed tasks, we apply a global penalty multiplier. Let $\bar{t}_{fail}$ be the average number of steps taken across all failed tasks, and let $S$ be the maximum steps allowed. The final WES is:

\begin{align*}
    \text{WES} &= \text{WES}^+ \cdot \left(1 - \frac{\bar{t}_{fail}}{S}\right)
\end{align*}

The penalty multiplier $(1 - \bar{t}_{fail}/S)$ ranges from 0 to 1. A multiplier of 1 means that the agent completes most tasks successfully or fails very quickly. This is by design, as the impracticality of inefficient agents will hamper their usage. A multiplier approaching 0 means the agent fails to complete most tasks \textit{and} is very inefficient, exhausting its full step budget on failures. This favors an agent that fails quickly over one that takes the full allotment of steps. Together, $\text{WES}^+$ and the penalty multiplier provide a clear picture of an agent's accuracy and efficiency. 

Notably, agentic systems may each have different values for $S$, which will result in different penalty multipliers. This is because $S$ acts as a cutoff point; if a system sets $S$ to a higher value, it may take more steps for a particular task. Thus, using each agent's value for $S$ when penalizing failures is the most fair policy.

\subsection{Results}

Table \ref{tab:full-results} presents the single-action and grouped-action WES scores for all the CUAs, together with their success rate as reported by OSWorld and the failure penalty multiplier. We segment the CUAs by observation type, as the OSWorld leaderboard does. For each CUA, we use its reported trajectory and our definition of grouped actions to determine which actions can be grouped.

The best-performing baselines on OSWorld also perform the best on both single-action and grouped-action WES. Relative ordering is also largely preserved. However, the absolute performance value is drastically reduced. Agent S2 w/ Gemini 2.5 holds the highest score on single-action WES (15.6\%) and grouped-action WES (9.6\%), a 2.7\x{} and 4.3\x{} reduction respectively from the comparative OSWorld score of 41.4\%. Intuitively, $\text{WES}^+$ represents the average number of \textit{extra} steps the agent takes, while the penalty multiplier captures how much of the step budget is wasted on failures. The performance when using multi-action trajectories for computing the expected number of steps is strictly worse across all baselines, as expected.

\subsection{Future Work}

Based on our study and benchmark results, we propose the following potential solutions to improve CUA efficiency:
\begin{itemize}
    \item \textbf{Action Grouping} - fusing multiple actions into a single step decreases the overall number of LLM calls, as described in Section \ref{sec:action-grouping}. This is challenging, as new content may emerge on the screen and invalidate future actions. Ideally, the planner model can also provide conditional actions that can be resolved by a grounding model downstream. 
    \item \textbf{Efficient Rollback} - being able to both recognize and rollback to an error-free state is critical, especially as agents can get caught in loops. We have provided a possible loop detection scheme in Section \ref{sec:cost-and-failure}. Once detected, a loop could be rolled back by prompting the agent to output both ``forwards" and ``backwards" actions.
    \item \textbf{Grounding Model Post-Training} - currently, grounding models provide coordinates from an entire screenshot. Post-training grounding models using reinforcement learning or traditional computer vision techniques, such as item segmentation, could improve locality.
    \item \textbf{History Compression} - naively passing the entire trajectory substantially increases the cost and latency of a request. Other techniques, such as using a sliding window, would make the cost of each step constant, but it may leave out important information from different steps in the trajectory. Compressing the history to a fixed window size while retaining important context would not only improve efficiency, but could also boost accuracy as the model will require fewer tokens.
    \item \textbf{Improvements in LLM Serving} - orthogonally, faster LLM serving techniques, such as prefix caching~\cite{sglang} and speculative decoding~\cite{leviathan2023fast, miao2024specinfer}, could reduce the per-step latency. Providers could offer a guaranteed prefix cache for computer-use and other multi-turn use cases would allow for larger context windows while keeping the per-step latency constant. With prefix caching, increase in reasoning tokens, and more complex action plans, the bottleneck shifts to decoding, where speculation could significantly improve latency. 
\end{itemize}

We leave the implementation of these solutions to future work.







%% file: tab_full_results_update.tex
\begin{table*}[!t]
\small
  \begin{center}
  \caption{Performance of SOTA CUAs on \sys\ with Single-Action and Grouped-Action Trajectories}
    \vspace{1em}
  \begin{tabular}{ccccccc}
    \hline
    Observation & Baseline (Max Steps) & Original (\%) & \textbf{Single-Action WES (\%)} & \textbf{Grouped-Action WES (\%)} \\
    \hline
    Screenshot (SS) & UI-TARS-1.5 (100) & 40.3 & 14.4 & 9.1 \\
    & Agent S2 w/ Gemini 2.5 (50) & \textbf{41.4} & \textbf{15.6} & \textbf{9.6} \\
    & InfantAgent (50) & 35.0 & 8.4 & 5.3 \\ 
    & Agent S2 w/ Claude 3.7 (50) & 34.5 & 6.6 & 4.4 \\ 
    & UI-TARS-1.5 7B (100) & 26.5 & 6.6 & 4.3 \\
    & UI-TARS-72B-DPO (50) & 23.7 & 11.0 & 7.4 \\
    \hline
    A11y Tree & GPT-4 (15) & \textbf{11.7} & \textbf{5.0} & \textbf{3.5} \\
    & GPT-4o (15) & 10.6 & 4.3 & 2.8 \\
    & Qwen-Max (15) & 6.3 & 2.6 & 1.5 \\
    & Gemini-Pro-1.5 (15) & 4.5 & 1.3 & 0.9 \\ 
    & Llama-3-70B (15) & 1.3 & 0.1 & 0.1 \\
    \hline
    SS + A11y Tree & GPT-4V (15) & \textbf{11.4} & 4.0 & 2.7 \\
    & GPT-4o (15) & 10.9 & \textbf{8.5} & \textbf{5.8} \\ 
    & Gemini-Pro-1.5 (15) & 4.6 & 0.8 & 0.5 \\
    \hline
    Set-of-Mark & GPT-4V (15) & \textbf{11.5} & \textbf{3.6} & \textbf{2.4} \\
    & Gemini-Pro Vision (15) & 0.8 & 0.1 & 0.1 \\
  \end{tabular}
  \label{tab:full-results}
  \end{center}
\end{table*}

%% file: conclude.tex
\section{Conclusion}
This paper performs the first study on the latency behavior of computer-use agents. We conduct a detailed analysis of Agent S2 and GTA1 on 39 OSWorld tasks. Our findings suggest LLM calls to be the major latency bottleneck and many steps could potentially be avoided without affecting the end results. We also conduct failure and cost analysis and find significant room for improvement when it comes to grounding models. Accordingly, we construct \sys, a manually-annotated version of OSWorld with human-determined task trajectories. We evaluate \sys\ on 16 CUAs and found that leading agents are extremely inefficient with 2.7$-$4.3\x{} longer trajectories than necessary. We believe this work and its open-sourced dataset will foster new research directions for computer-use agents.





